%% file: ms.tex
\documentclass[10pt,twocolumn,letterpaper]{article}

\usepackage{cvpr}
\usepackage{times}
\usepackage{epsfig}
\usepackage{graphicx}
\usepackage{amsmath}
\usepackage{amssymb}
\usepackage{multirow}
\usepackage{subcaption}

\cvprfinalcopy % *** Uncomment this line for the final submission

\def\cvprPaperID{4254} % *** Enter the CVPR Paper ID here

\ifcvprfinal\fi
\begin{document}

%%%%%%%%% TITLE
\title{Eye In-Painting with Exemplar Generative Adversarial Networks}

\author{Brian Dolhansky, Cristian Canton Ferrer\\
Facebook Inc.\\
1 Hacker Way, Menlo Park (CA), USA\\
{\tt\small\{bdol, ccanton\}@fb.com}
}

\twocolumn[{%
\renewcommand\twocolumn[1][]{#1}%
\maketitle
\begin{center}
	\vspace*{-5mm}
    \centering
    \includegraphics[width=1.0\textwidth]{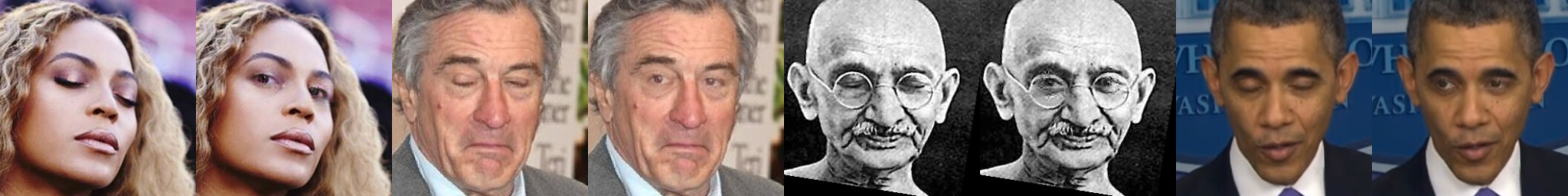}
    \vspace*{2mm}
\end{center}%
}]

\begin{abstract}

\par This paper introduces a novel approach to in-painting where the identity of the object to remove or change is preserved and accounted for at inference time: Exemplar GANs (ExGANs). ExGANs are a type of conditional GAN that utilize exemplar information to produce high-quality, personalized in-painting results. We propose using exemplar information in the form of a reference image of the region to in-paint, or a perceptual code describing that object. Unlike previous conditional GAN formulations, this extra information can be inserted at multiple points within the adversarial network, thus increasing its descriptive power. We show that ExGANs can produce photo-realistic personalized in-painting results that are both perceptually and semantically plausible by applying them to the task of closed-to-open eye in-painting in natural pictures. A new benchmark dataset is also introduced for the task of eye in-painting for future comparisons.
\end{abstract}

%%%%%%%%% BODY TEXT
\input{intro.tex}
\input{related.tex}
\input{model.tex}
\input{experiments.tex}

\input{results.tex}
\input{conclusions.tex}

\clearpage
\newpage
{\small
\bibliographystyle{ieee}
\bibliography{refs}
}

\end{document}

%% file: intro.tex
\vspace*{-3mm}
\section{Introduction}

\par Every day, around 300M pictures are captured and shared in social networks with a large percentage of them featuring people-centric content. There is little doubt that realistic face retouching and beautification algorithms are a growing research topic within the computer vision and machine learning communities. Some examples include red-eye fixing~\cite{Yoo2009} and blemish removal~\cite{Brand2008}, where patch matching and Poisson blending have been used to create plausible-looking results~\cite{Yang2011}. Full manipulation of the face appearance like beautification~\cite{Leyvand2006}, attribute transferral~\cite{Vlasic2005}, face frontalization~\cite{Sagonas2015} or synthetic make-up~\cite{Guo2009}, are also becoming very popular. However, humans are very sensitive to small errors in facial structure, specially if those faces are our own or are well-known to us~\cite{Sinha2006}; moreover, the so-called "uncanny valley"~\cite{Mori1970} is a difficult impediment to cross when manipulating facial features.

%% These next two paragraphs might be combined in a single one.
\par Recently, deep convolutional networks (DNNs) have produced high-quality results when in-painting missing regions of pictures showing natural scenery~\cite{Iizuka2017}. For the particular problem of facial transformations, they learn not only to preserve features such global lighting and skin tone (which patch-like and blending techniques can also potentially preserve), but can also encode some notion of semantic plausibility. Given a training set of sufficient size, the network will learn what a human face "should" look like~\cite{Karras2017}, and will in-paint accordingly, while preserving the overall structure of the face image.

\par In this paper, we will focus on the particular problem of \textbf{eye in-painting}. While DNNs can produce semantically-plausible, realistic-looking results, most deep techniques do not preserve the identity of the person in a photograph. For instance, a DNN could learn to open a pair of closed eyes, but there is no guarantee encoded in the model itself that the new eyes will correspond to the original person's specific ocular structure. Instead, DNNs will insert a pair of eyes that correspond to similar faces in the training set, leading to undesirable and biased results; if a person has some distinguishing feature (such as an uncommon eye shape), this will not be reflected in the generated part.

\par Generative adversarial networks (GANs) are a specific type of deep network that contain a learnable adversarial loss function represented by a discriminator network~\cite{Goodfellow2014}. GANs have been successfully used to generate faces from scratch~\cite{Karras2017}, or to in-paint missing regions of a face~\cite{Iizuka2017}. They are particularly well-suited to general facial manipulation tasks, as the discriminator uses images of real faces to guide the generator network into producing samples that appear to arise from the given ground-truth data distribution. One GAN variation, conditional-GANs (or cGANs), can constrain the generator with extra information, and have been used to generate images based on user generated tags~\cite{Mirza2014}. However, the type of personalization described above (especially for humans) has not been previously considered within the GAN literature.

\par This paper extends the idea of using extra conditional information and introduces Exemplar GANs (ExGANs), a type of a cGAN where the extra information corresponds directly to some identifying traits of the entity of interest. Furthermore, we assume that this extra information (or "exemplar") is available at inference time. We believe that this is a reasonable assumption since multiple images of the same objects are readily available. Exemplar data is not restricted to raw images, and we prove that a perceptually-coded version of an object can also be used as an exemplar.

\par The motivation for the use of exemplar data is twofold. First, by utilizing extra information, ExGANs do not have to hallucinate textures or structure from scratch, but will still retain the semantics of the original image. Second, output images are automatically \emph{personalized}. For instance, to in-paint a pair of eyes, the generator can use another exemplar instance of those eyes to ensure the identity is retained. 

\par Finally, ExGANs differ from the original formulation of a cGAN in that the extra information can be used in multiple places; either as a form of perceptual loss, or as a hint to the generator or the discriminator. We propose a general framework for incorporating this extra exemplar information. As a direct application, we show that using guided examples when training GANs to perform eye in-painting produces photo-realistic, identity-preserving results.

%% file: related.tex
\section{Related Work}

%\par 

\begin{figure}[t]
  \centering
  \begin{subfigure}[b]{0.12\textwidth}
  	\includegraphics[width=\textwidth]{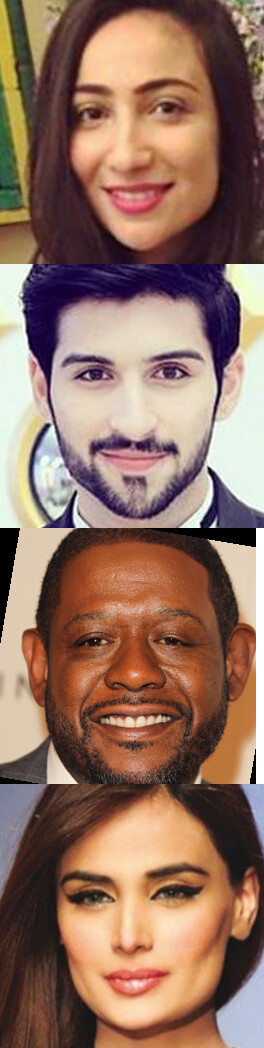}
    \caption{}
  \end{subfigure}%
  \begin{subfigure}[b]{0.12\textwidth}
  	\includegraphics[width=\textwidth]{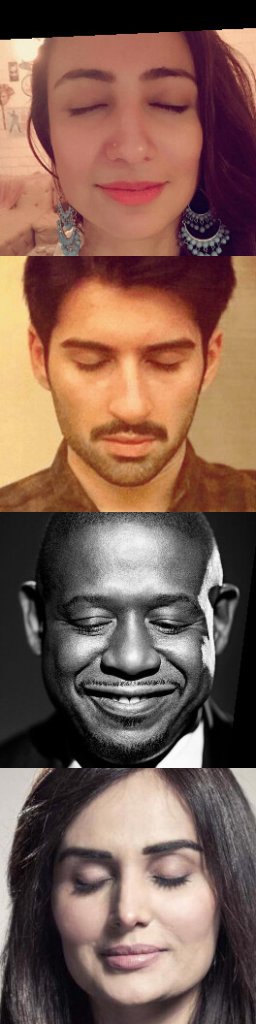}
    \caption{}
  \end{subfigure}%
  \begin{subfigure}[b]{0.12\textwidth}
  	\includegraphics[width=\textwidth]{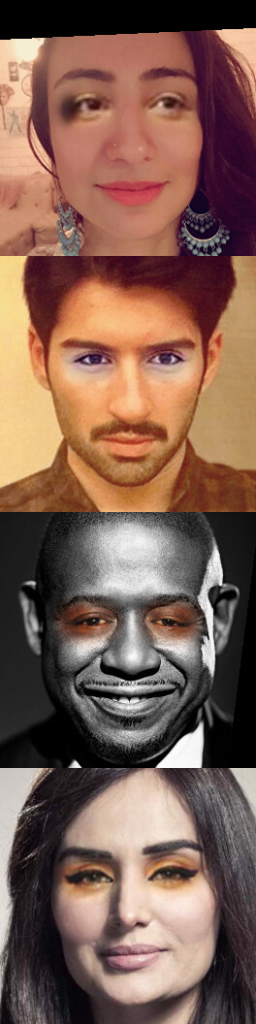}
    \caption{}
  \end{subfigure}%
  \begin{subfigure}[b]{0.12\textwidth}
  	\includegraphics[width=\textwidth]{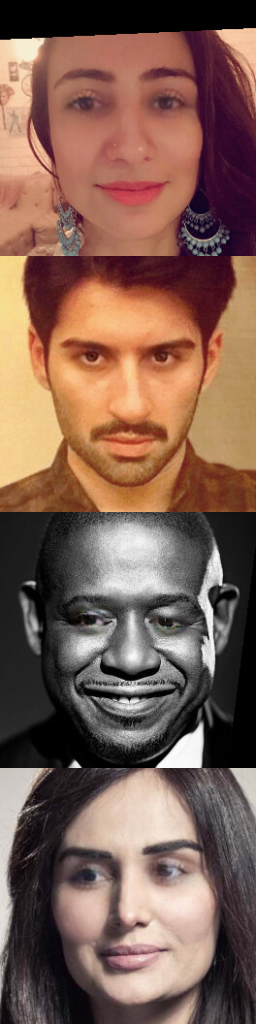}
    \caption{}
  \end{subfigure}%
  \caption{Comparison between the commercial state of the art eye opening algorithm in Adobe Photoshop Elements~\cite{Adobe} (c) and the proposed ExGAN technique (d). The exemplar and original images are shown in (a) and (b), respectively.\vspace*{-3mm}}
  \label{fig:ours_v_patch}
\end{figure}

%Our method is able to take pose and global lighting into account.

\par Previous approaches to opening closed eyes in photographs have generally used example photos, such as a burst of photographs of a subject in a similar pose and lighting conditions~\cite{Agarwala2004}, and produced final results with a mixture of patch matching~\cite{Barnes2009} and blending~\cite{Perez2003}. However, this technique does not take full advantage of semantic or structural information in the image, such as global illumination or the pose of the subject. Small variations in lighting or an incorrect gaze direction produce uncanny results, as seen in Fig.~\ref{fig:ours_v_patch}. 

Besides classic computer vision techniques, recent research has focused on using deep convolutional networks to perform a variety of facial transformations. Specifically within this body of work, the applications of GANs~\cite{Goodfellow2014} to faces are numerous~\cite{Hassner,Li2017,Yeh2016}. Many GANs are able to generate photo-realistic faces from a single low-dimensional vector~\cite{Kingma2014}, pushing results out of the uncanny valley and into the realm of reality. Fader networks~\cite{Lample2017} expand on this idea by training in such a way as to make each element of the low-dimensional noise vector correspond to a specific face attribute, such as beards or glasses. By directly manipulating these elements, parts can be transferred or changed on demand, including opening or closing a mouth or changing a frown into a smile. However, identity is not preserved with this technique.

In-painting has been studied extensively, both with and without deep networks~\cite{Bertalmio2000, Yang2016}. Exemplar In-painting~\cite{Bertalmio2003} is an iterative algorithm that decomposes an image into its structural and textured components, and holes are reconstructed with a combination of in-painting and texture synthesis. This technique has been used to remove large objects from images~\cite{Criminisi2003,Criminisi2004}, and its effectiveness has been compared to deep methods, where it is shown that Exemplar In-painting struggles with complex or structured in-painting~\cite{Yoshida}. More recently, cGANs~\cite{Mirza2014} have been used with success when in-painting natural images, by using extra information such as the remaining portions of an image to in-paint. 

The generator network in a GAN learns to fill in missing regions of an image, and the discriminator network learns to judge the difference between in-painted and real images, and can take advantage of discontinuities between the in-painted and original regions. This forces the generator to produce in-painted results that smoothly transition into the original photograph, directly sidestepping the need for any pixel blending. Besides the general case of in-painting scenes, GANs have also been used to in-paint regions of a face~\cite{Gauthier2014}. At inference time, these GANs must rely on information that is present only in the training set, and are incapable of personalized face in-paintings, unless that particular face also exists in the training set.

Finally, of particular relevance is the work on multi-view face synthesis, and specifically the approaches that attempt to preserve the identity of a given face. In the face identification regime, pose invariance is particularly important, and previous work has focused on developing various identity-preserving objectives. One approach inputs a set of training images containing multiple views of the same face~\cite{Zhu2014}, and attempts to generate similar views of a separate input face at inference time. An identity-preserving loss is proposed in~\cite{Huang2017}, which uses a perceptual distance of two faces on the manifold of the DNN outlined in \cite{Zhu2013} as an objective to be minimized. However, unlike the aforementioned approaches, we make the assumption that a reference image will be available at inference time. Like these approaches, a perceptual code can be generated from the reference face, but we also propose that just providing the generator the raw reference image can also help with identity preservation.

%% file: model.tex
\section{Exemplar GANs for in-painting}
\par Instead of relying on the network to generate images based only on data seen in the training set, we introduce ExGANs, which use a second source of related information to guide the generator as it creates an image. As more datasets are developed and more images are made available online, it is reasonable to assume that a second image of a particular object exists at inference time. For example, when in-painting a face, the reference information could be a second image of the same person taken at a different time or in a different pose. However, instead of directly using the exemplar information to produce an image (such as using nearby pixels for texture synthesis, or by copying pixels directly from a second photograph), the network learns how to incorporate this information as a semantic guide to produce perceptually-plausible results. Consequently, the GAN learns to utilize the reference data while still retaining the characteristics of the original photograph.

%\par A desirable side effect of exemplar in-painting is the automatic personalization of results. Specifically, when in-painting eyes, it is not enough to in-paint "a" pair of eyes onto a face; painting "the" pair of eyes belonging to that face is necessary to retain the original identity. Although not covered in this work, in other cases it may be required to generate images of non-human objects that have some uniquely identifying aspect. For instance, pictures of the New York City might be used to generate alternate angles from a set of reference photos, or the reference photos may be used to in-paint and remove undesired objects from the foreground of a photograph of the skyline.

\par We propose two separate approaches to ExGAN in-painting. The first is reference-based in-painting, in which a reference image $\mathbf{r}_i$ is used in the generator as a guide, or in the discriminator as additional information when determining if the generated image is real or fake. The second approach is code-based in-painting, where a perceptual code $\mathbf{c}_i$ is created for the entity of interest. For eye in-painting, this code stores a compressed version of a person's eyes in a vector $\mathbf{c}_i \in \mathbb{R}^N$, which can also be used in several different places within the generative and discriminator networks.

\par Formally, both approaches are defined as a two-player minimax game, where each objective is conditioned on extra information, similar to \cite{Mirza2014}. This extra information can be the original image with patches removed, $\mathbf{r}_i$, or $\mathbf{c}_i$, or some combination of these. An additional content loss term can be added to this objective. The framework is general, and can potentially be applied to tasks other than in-painting.

\subsection{Reference image in-painting}

\par Assume that for each image in the training set $\mathbf{x}_i$, there exists a corresponding reference image $\mathbf{r}_i$. Therefore the training set $X$ is defined as a set of tuples ${X=\{(\mathbf{x}_1, \mathbf{r}_1),\ \ldots,\ (\mathbf{x}_n, \mathbf{r}_n)\}}$. For eye in-painting, $\mathbf{r}_i$ is an image of the same person in $\mathbf{x}_i$, but potentially taken in a different pose. Patches are removed from $\mathbf{x}_i$ to produce $\mathbf{z}_i$, and the learning objective is defined as:
\begin{equation}
\begin{split}
	\min_G & \max_D\ V(D, G) =\ \mathbb{E}_{
    	\mathbf{x}_i, \mathbf{r}_i\sim p_\text{data}(\mathbf{x}, \mathbf{r})
    } 
    [ \log D(\mathbf{x}_i, \mathbf{r}_i) ]\ + \\
    & \mathbb{E}_{
    	\mathbf{r_i} \sim p_\mathbf{r}, G(\cdot) \sim p_\mathbf{z}
    } [\log 1 - D(G(\mathbf{z}_i, \mathbf{r}_i))]\ + \\
    & ||G(\mathbf{z}_i, \mathbf{r}_i) - \mathbf{x}_i ||_1
\end{split}
\label{eq:ref_objective}
\end{equation}
This objective is similar to the standard GAN formulation in \cite{Goodfellow2014}, but both the generator and discriminator can take an example as input.

For better generalization, a set of reference images $R_i$ corresponding to a given $\mathbf{x}_i$ can also be utilized, which expands the training set to the set of tuples comprised of the Cartesian product between each image-to-be-in-painted and its reference image set, $X=\{\mathbf{x}_1\times R_1,\ \ldots,\ \mathbf{x}_n\times R_n\}$.

\subsection{Code in-painting}
\par For code-based in-painting, and for datasets where the number of pixels in each image is $|I|$, assume that there exists a compressing function $C(\mathbf{r}): \mathbb{R}^{|I|} \rightarrow \mathbb{R}^N$, where $N \ll |I|$. Then, for each image to be in-painted $\mathbf{z}_i$ and its corresponding reference image $\mathbf{r}_i$, a code $\mathbf{c}_i = C(\mathbf{r}_i)$ is generated using a $\mathbf{r}_i$. Given the codified exemplar information, we define the adversarial objective as:

\begin{equation}
\begin{split}
	\min_G & \max_D\ V(D, G) =\ \mathbb{E}_{
    	\mathbf{x}_i, \mathbf{c}_i\sim p_\text{data}(\mathbf{x}, \mathbf{c})
    } 
    [ \log D(\mathbf{x}_i, \mathbf{c}_i) ]\ + \\
    & \mathbb{E}_{
    	\mathbf{c_i} \sim p_\mathbf{c}, G(\cdot) \sim p_\mathbf{z}
    } [\log 1 - D(G(\mathbf{z}_i, \mathbf{c}_i))]\ + \\
    & ||G(\mathbf{z}_i, \mathbf{c}_i) - \mathbf{x}_i||_1 + 
      ||C(G(\mathbf{z}_i, \mathbf{c}_i) - \mathbf{c}_i ||_2
\end{split}
\label{eq:code_objective}
\end{equation}

\par The compressing function can be a deterministic function, an auto-encoder, or a general deep network that projects an example onto some manifold. The final term in Eq.~\ref{eq:code_objective} is an optional loss that measures the distance of the generated image $G(\mathbf{z}_i, \mathbf{c}_i)$ to the original reference image $\mathbf{r}_i$ in a perceptual space. For a deep network, this corresponds to measuring the distance between the generated and reference images on a low-dimensional manifold. Note that if the generator $G$ is originally fully-convolutional, but takes $\mathbf{c}_i$ as input, its architecture must be modified to handle an arbitrary scalar vector. 

\begin{figure}[!t]
\begin{center}
\includegraphics[width=\linewidth]{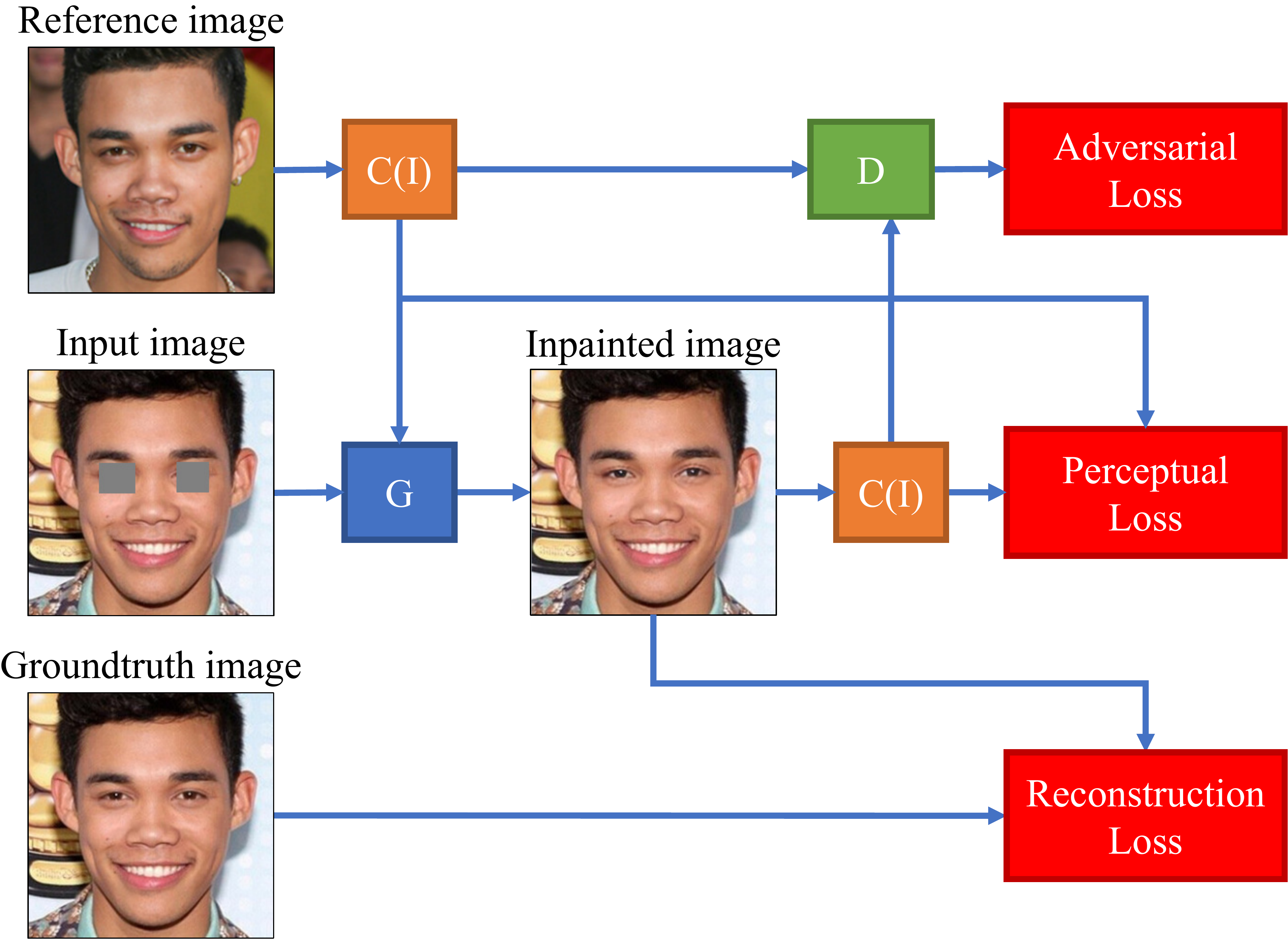}
\end{center}
   \caption{General architecture of an Exemplar GAN. The overall training flow can be summarized as (1) mark the eyes from the input image; (2) in-paint the image with the reference image or code as a guide; (3) compute the gradient of the generator's parameters via the content/reconstruction loss between the input image and the in-painted image; (4) compute the gradient of the discriminator's parameters with the in-painted image, another real, ground truth image, and the reference image or code; (5) backpropagate the discriminator error through the generator. Optionally, (6) the generator's parameters can also be updated with a perceptual loss. For reference-based Exemplar GANs, the compressing functions $C(I)$ are the identity function.\vspace*{-3mm}}
\label{fig:scheme}
\end{figure}

\subsection{Model architecture}

\par The overall layout for both code- and reference-based ExGANs is depicted in Fig.~\ref{fig:scheme}. For most experiments, we used a standard convolutional generator, with a bottleneck region containing dilated convolutions, similar to the generator proposed in~\cite{Iizuka2017}, but with a smaller channel count in the interior layers of the network, as generating eyes is a more restricted domain than general in-painting. The input to the generator is an RGB image with the portions to in-paint removed, stacked with a one-channel binary mask indicating which regions to fill. The generator could take an additional four channels: the RGB values of the reference image (with no missing regions), and another 1-channel mask indicating the eye locations. All eye locations are detected prior to training and stored with the dataset.

The discriminator is similar to the global/local discriminator developed in \cite{Iizuka2017}. This discriminator processes both the whole face (the global region) and a zoomed-in portion of the image that contains the eyes (the local region). Having a global adversarial loss enforces overall semantic consistency, while the local adversarial loss ensures detail and sharpness in the generated output. The outputs of the global and local convolutional branches are concatenated and passed through a final sigmoid. An additional global branch is added to the discriminator if a reference image is being used as input to $D$. In this case, the outputs of all three branches are concatenated.

Next, because of the possibility that the generator network could take $\mathbf{c}_i$ as input in Eq.~\ref{eq:code_objective}, we tested an alternative architecture to the fully-convolutional generator. This generator uses an encoder-decoder architecture, with 4 downsampling and upsampling layers, and with a 256 dimensional fully-connected bottleneck layer. The bottleneck is concatenated with the eye code, resulting in an overall dimensionality of 512 at the output of the bottleneck. The eye code can also be used in the perceptual loss term of Eq.~\ref{eq:code_objective}. Furthermore, the code can be appended to the penultimate, fixed-size output of the discriminator. Because the 256 dimensions of the code is much larger than the two outputs of the original discriminator, we experimented with feeding the global and local outputs and the code through a small two-layer fully-connected network before the final sigmoid in order to automatically learn the best weighting between the code and the convolutional discriminator. For the remainder of this paper, any reference to code-based ExGANs used this architecture.

To generate $\mathbf{c}_i$, we trained a separate auto-encoder for the compressing function $C$, but with a non-standard architecture. During training of $C$, the encoder took a single eye as input, but the decoder portion of the autoencoder split into a left and right branch with separate targets for both the left and right eyes. This forces the encoder to learn not to duplicate features common to both eyes (such as eye color), but to also encode distinguishing features (such as eye shape). In general, each eye was encoded with a 128 dimensional float vector, and these codes were combined to form a 256 dimensional eye code.

Unless otherwise specified, ELU~\cite{Clevert2015} activations were used after all convolution layers. We also implemented one-sided label smoothing~\cite{Salimans2016} with probability 0.05. A full listing of model architectures is given in the supplemental material.

%% file: experiments.tex
\section{Experiment setup}

\par ExGANs require a dataset that contain pairs of images for each object, but these types of datasets are not as common. We observed that we require a large number of unique identities for sufficient generalization. High resolution images taken in a variety of environments and lighting conditions permits an ExGAN to be able to in-paint eyes for a wide variety of input photographs. In addition, including images without distractors and in non-extreme poses improved the quality and sharpness of the generated eyes. We were not able to utilize the CelebA~\cite{Liu2015} dataset as it only contains 10K unique identities. Furthermore, the photos in CelebA were usually taken in unnatural environments, such as red carpet photographs or movie premieres. The MegaFace~\cite{Nech2017} dataset provides a more suitable training set, but many images do not contain human faces and those that do include faces with sunglasses or in extreme poses. We desired a finer-grained control over certain aspects of our dataset, such as ensuring that each image group contained the same individual with high confidence and that there were no distracting objects on the face.

\par In order to circumvent the limitations of pre-existing datasets, we developed an internal training set of roughly 2~million 2D-aligned images of around 200K individuals. For each individual, at least 3 images were present in the dataset. Every image in the training set contained a person with their eyes opened to force the network to only learn to in-paint open eyes. 

\begin{figure}[t]
\begin{center}
   \includegraphics[width=\linewidth]{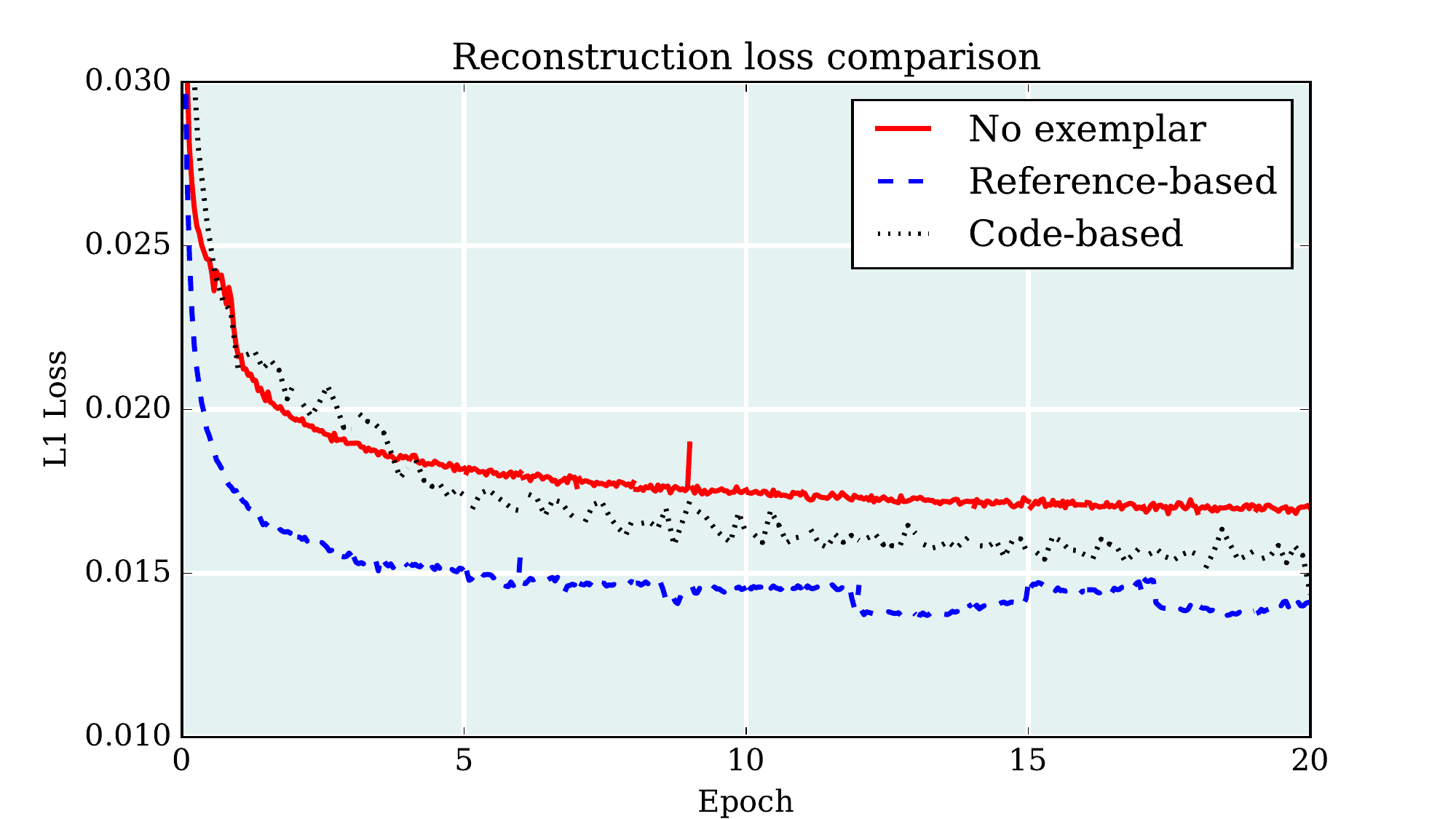}
\end{center}
   \caption{Training reconstruction loss comparison between non-exemplar and Exemplar GANs.\vspace*{-4.5mm}}
\label{fig:model_rec_loss}
\end{figure}

\par For external replication purposes we developed an eye in-painting benchmark from high-quality images of celebrities scraped from the web. It contains around 17K individual identities and a total of 100K images, with at least 3 photographs of each celebrity. An additional, we created a publicly-available benchmark called Celeb-ID\footnote{\texttt{https://bdol.github.io/exemplar\_gans}}. Note that the network was \textit{only} trained on our internal dataset, thereby making it impossible for our network to overfit to any images shown in this paper as we did not use any celebrity images during training. 

\par During a training epoch, for each individual, one image was inpainted and a second, different image was used either as an example for the network, or used to generate $\mathbf{c}_i$. The generator and discriminator were trained for a variable number of epochs (depending on the dataset size), but in general, we stopped training after the network saw 2M image pairs. Each training run took about 3 days to train on two Tesla M40 GPUs. 

\par Each objective was optimized using Adam~\cite{Kingma2014} and with parameters $\beta_1, \beta_2 = 0.9, 0.999$. In order to emphasize the viability of exemplar in-painting, only L1 distance was used for the reconstruction loss, and binary cross-entropy was used for the adversarial loss. We did not use any learning rate tricks such as gradient regularization~\cite{Arjovsky2017} or a control-theory approach~\cite{Kingma2014}. The hyperparameters swept included all learning rates, the relative weight of the discriminator network, the weight assigned to the perceptual loss, and at which points in the network to use a reference image or the eye code. A full table of various results for all experiments is given in the supplemental material.

%% file: results.tex
\section{Results}
In order to best judge the effects of both code- and reference-based ExGANs, we avoided mixing codes and reference images in a single network. Throughout this section, we compare and contrast the results of three models: (1) a non-exemplar GAN, with an architecture identical to the global/local adversarial net of ~\cite{Iizuka2017}, with the only difference being a smaller channel count in the generator, (2) our best reference image Exemplar GAN and (3) our best code-based Exemplar GAN. We tried multiple other GAN architectures, but the model introduced in ~\cite{Iizuka2017} produced the best non-exemplar results. Note that each GAN in this comparison has the same base architecture and hyperparemters, with the exception of the code-based GAN, which uses an encoder-decoder style generator. Interestingly, the same learning rate could be used for both types of generators, most likely because they had a similar number of parameters and depth. In this particular setup, the perceptual loss had little overall effect on the final quality of the generator output; instead, better results were generated when using the eye code directly in the generator itself. 

In Fig.~\ref{fig:model_rec_loss}, we show the effect of exemplars on the overall reconstruction loss. With the addition of eye codes, the content loss of the non-exemplar GAN is decreased by 8\%, while adding reference images decreased the L1 loss by 17\%. During training, models that had a low overall content loss and at least a decreasing adversarial loss tended to produce the best results. Training runs with a learning rate of 1e-4 for both the generator and discriminator resulted in the most well-behaved loss decrease over time. However, for eye in-painting, we determined that the content loss was not entirely representative of the final perceptual quality, an issue discussed further in Section~\ref{sec:ab_test}.

\begin{figure}[t]
\begin{center}
  \begin{subfigure}[b]{0.104\textwidth}
  	\includegraphics[width=\textwidth]{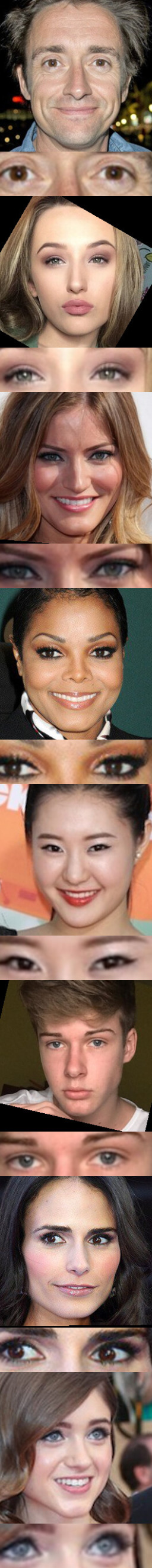}
    \caption{}
  \end{subfigure}%
  \begin{subfigure}[b]{0.104\textwidth}
  	\includegraphics[width=\textwidth]{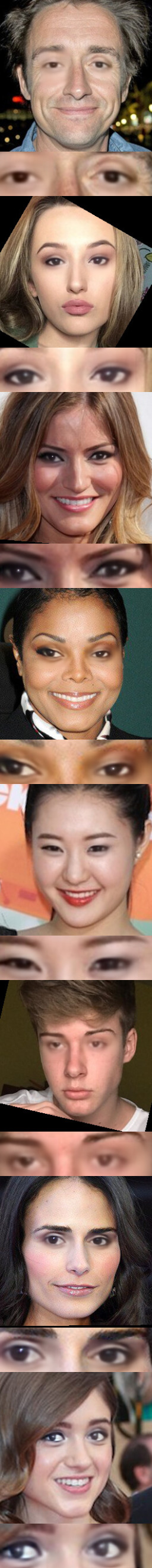}
    \caption{}
  \end{subfigure}%
  \begin{subfigure}[b]{0.104\textwidth}
  	\includegraphics[width=\textwidth]{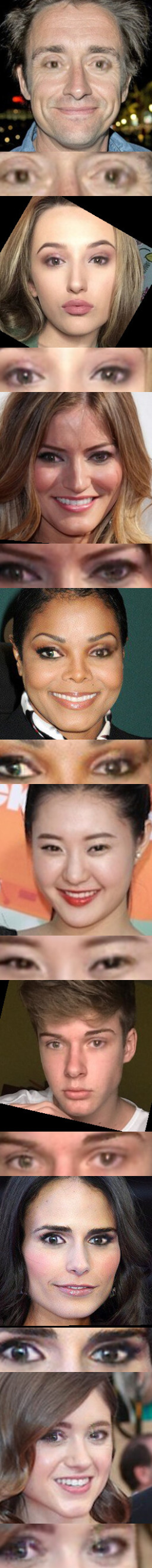}
    \caption{}
  \end{subfigure}%
  \begin{subfigure}[b]{0.104\textwidth}
  	\includegraphics[width=\textwidth]{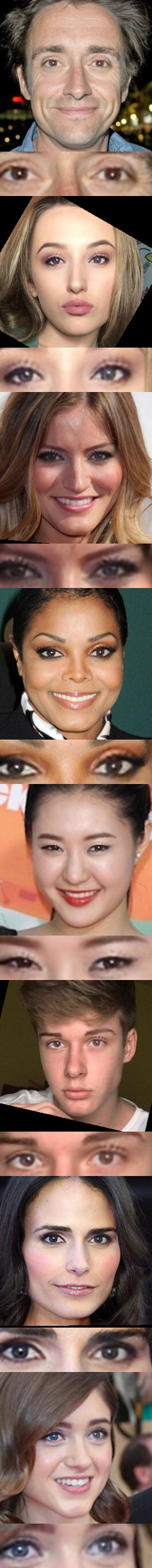}
    \caption{}
  \end{subfigure}%
\end{center}
   \vspace*{-6mm}
   \caption{Comparison between (a) ground truth, (b) non-exemplar and (c, d) exemplar-based results. An ExGAN that uses a reference image in the generator and discriminator is shown in column (c), and an ExGAN that uses a code is shown in column (d).}
\label{fig:model_compare}
\end{figure}

Next, in Fig.~\ref{fig:model_compare}, we compare the perceptual results generated by exemplar and non-exemplar GANs. As is evident in the figure, each of the ExGANs produce superior qualitative results, with the code-based exemplar model resulting in the most convincing and personalized in-paintings.

\begin{figure}[t]
  \centering
  \begin{subfigure}[b]{0.066\textwidth}
  	\includegraphics[width=\textwidth]{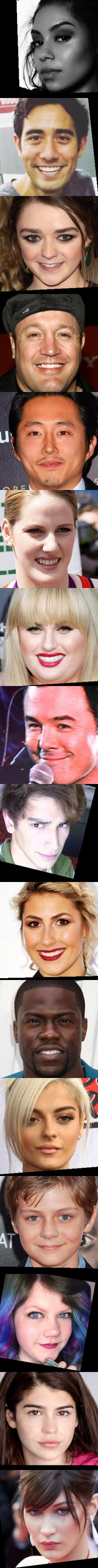}
    \caption{}
  \end{subfigure}%
  \begin{subfigure}[b]{0.066\textwidth}
  	\includegraphics[width=\textwidth]{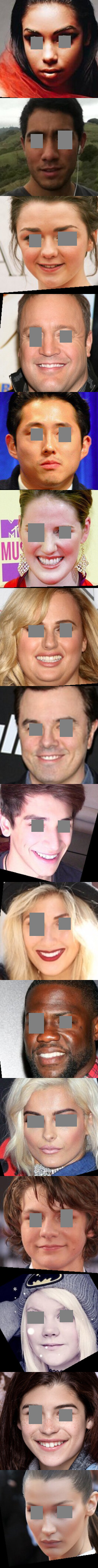}
    \caption{}
  \end{subfigure}%
  \begin{subfigure}[b]{0.066\textwidth}
  	\includegraphics[width=\textwidth]{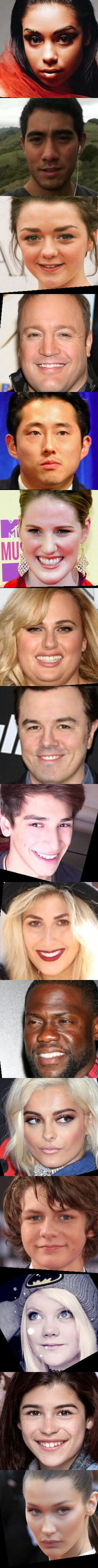}
    \caption{}
  \end{subfigure}%
  \begin{subfigure}[b]{0.066\textwidth}
  	\includegraphics[width=\textwidth]{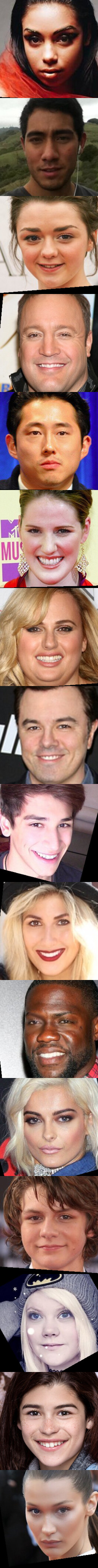}
    \caption{}
  \end{subfigure}%
  \begin{subfigure}[b]{0.066\textwidth}
  	\includegraphics[width=\textwidth]{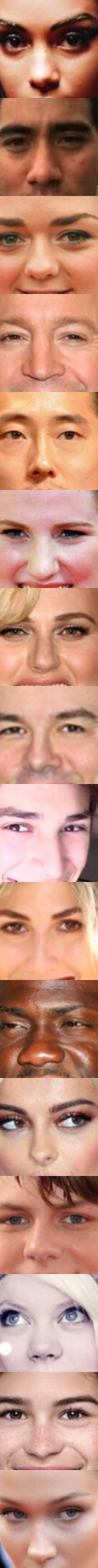}
    \caption{}
  \end{subfigure}%
  \begin{subfigure}[b]{0.066\textwidth}
  	\includegraphics[width=\textwidth]{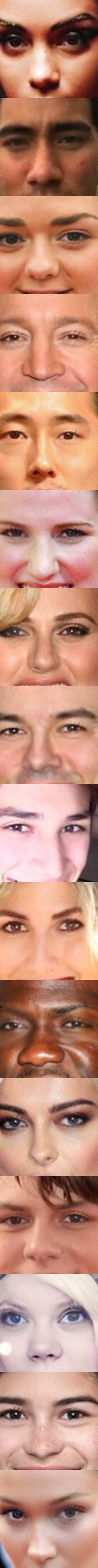}
    \caption{}
  \end{subfigure}%
   \caption{Results generated with a code-based Exemplar GAN. Columns represent: (a) reference image, (b) image to in-paint, (c) ground-truth global image, (d) in-painted global image, (e) ground-truth local image, (f) in-painted local image.}
\label{fig:celeb_high_q}
\end{figure}

\begin{figure}[!t]
\begin{center}
\includegraphics[width=\linewidth]{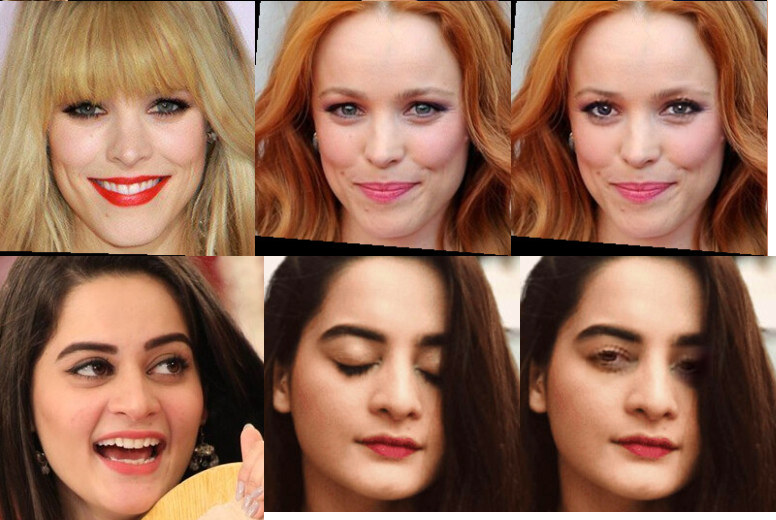}
\end{center}
   \caption{Failure cases of our models include not fully preserving the iris color (top row) or not preserving the shape (bottom row), especially if the face to in-paint has one occluded eye.}
\label{fig:failure_cases}
\end{figure}

\begin{figure}[t]
  \centering
  \begin{subfigure}[b]{0.15\textwidth}
  	\includegraphics[width=\textwidth]{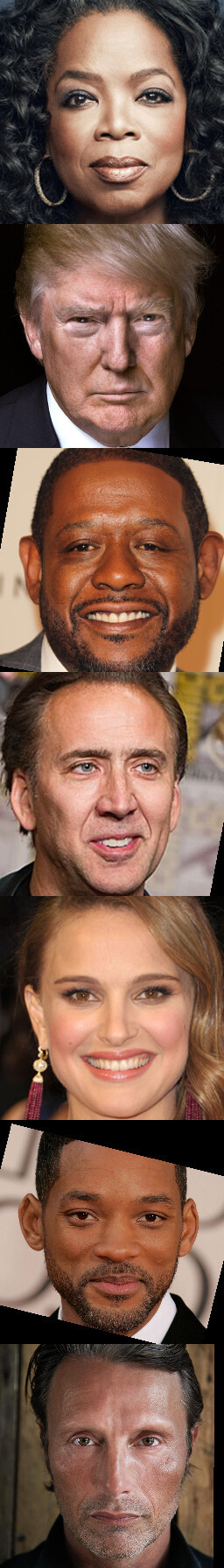}
    \caption{}
  \end{subfigure}%
  \begin{subfigure}[b]{0.15\textwidth}
  	\includegraphics[width=\textwidth]{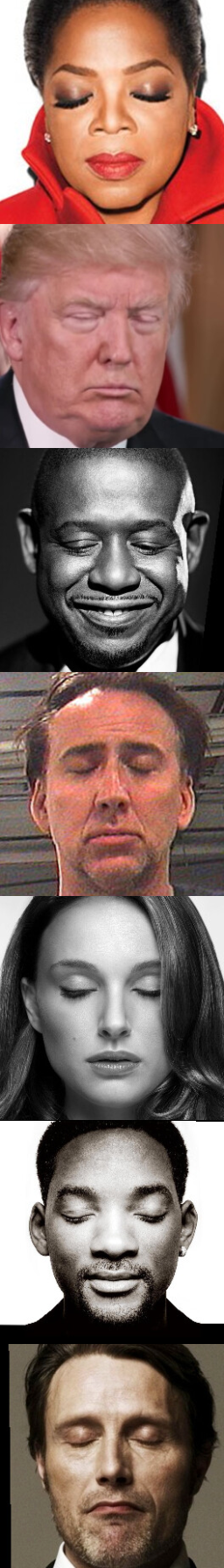}
    \caption{}
  \end{subfigure}%
  \begin{subfigure}[b]{0.15\textwidth}
  	\includegraphics[width=\textwidth]{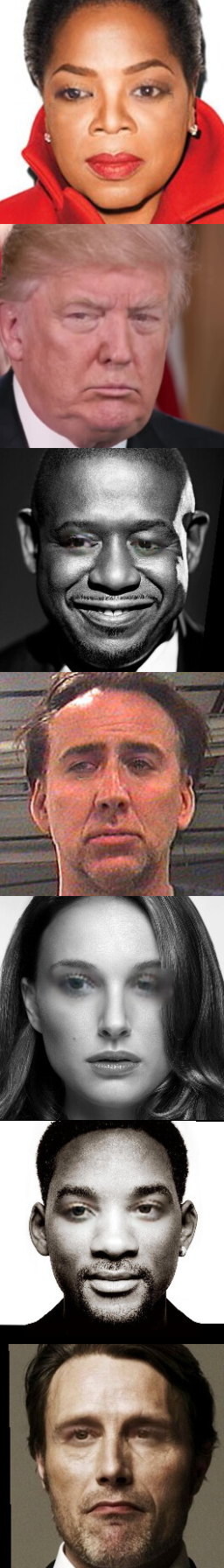}
    \caption{}
  \end{subfigure}%
  \caption{Additional closed-eye-opening results generated with a reference-based Exemplar GAN. Column (a) is the reference image, and column (c) is the in-painted version of the images in column (b) generated with an Exemplar GAN.\vspace*{3cm}}
  \label{fig:eyes_open}
\end{figure}

Finally, in Figs.~\ref{fig:celeb_high_q} and~\ref{fig:eyes_open}, we show additional qualitative results on the celebrity validation set, generated by an ExGAN that uses a code-based exemplar in both the generator and discriminator with no perceptual loss. Both the local and global in-painted images are shown along with the reference image used for in-painting. It is evident that the network matches the original eye shape and accounts for pose and lighting conditions in the in-painted image. In some cases, such as in Fig.~\ref{fig:failure_cases}, the network did not match the iris color exactly, most likely because a mismatch in the eye shape would incur a higher content or adversarial loss. We describe some solutions to this problem in Section~\ref{sec:conclusions}.

\subsection{Content loss vs. perceptual loss} \label{sec:ab_test}

\begin{figure}[t]
  \centering
  \begin{subfigure}[b]{0.15\textwidth}
  	\includegraphics[width=\textwidth]{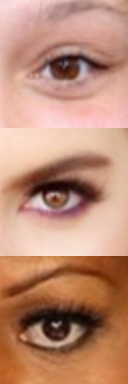}
    \caption{}
  \end{subfigure}%
  \begin{subfigure}[b]{0.15\textwidth}
  	\includegraphics[width=\textwidth]{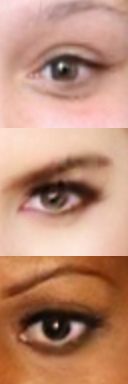}
    \caption{}
  \end{subfigure}%
  \begin{subfigure}[b]{0.15\textwidth}
  	\includegraphics[width=\textwidth]{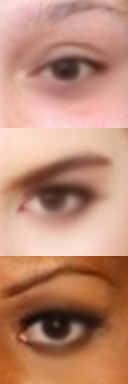}
    \caption{}
  \end{subfigure}%
  \caption{Comparison between (a) the ground-truth image (b) results from a model trained to epoch 10, with L1 loss 0.01457 and (c) results from a model at epoch 18, with L1 loss 0.01386. Despite the lower content loss, the model trained for longer produces blurrier results. The FID score is a better metric of perceptual quality; in this model the FID score at epoch 10 is 7.67, while at epoch 18 it is 10.55.\vspace*{-4mm}}
  \label{fig:loss_compare}
\end{figure}

In general, the content or adversarial losses were not one-to-one proxies for perceptual quality, as discussed in \cite{Ledig2016}. In many cases, a network with a low content loss produced training results that looked perceptually worse than another network with a slightly higher content loss. As an example, refer to Fig.~\ref{fig:loss_compare}, which includes the output of the same network for different values of the L1 losses. Although it may be that this effect is simply an example of overfitting, we also observed poor results for lower loss values on the \textit{training} set. This observation justifies the fact that perceived quality and raw pixel difference are only correlated up to a certain point. In order to combat this effect, we stopped training early as a form of regularization. 

In addition, we measured several perceptual metrics over the course of each model's training run, including MS-SSIM~\cite{Wang2003}, inception score~\cite{Salimans2016}, and FID score~\cite{Heusel2017}. Neither the MS-SSIM score nor the inception score correlated strongly with perceptual quality. We believe that the inception score specifically did not correlate as it is based on scores from the interior layers of GoogLeNet~\cite{Szegedy2014}, a net trained for image classification. As all generated images belong to the same class, the network activations did not vary enough with the fine-grained details around an eye.

\begin{table}
\begin{center}
\begin{tabular}{| l | c c c c|}
\hline
Model & L1 & MS-SSIM & Inception & FID \\
\hline
\multicolumn{5}{|c|}{Internal benchmark}\\
\hline
Non-exemplar & 0.018 & 5.05E-2 & 3.96 & 11.27\\
Reference & 0.014 & 3.97E-2 & 3.82 & 7.67\\
Code & 0.015 & 4.15E-2 & 3.94 & 8.49\\
\hline
\multicolumn{5}{|c|}{Celeb-ID}\\
\hline
Non-exemplar & 7.36E-3 & 8.44E-3 & 3.72 & 15.30\\
Reference & 7.15E-3 & 7.97E-3 & 3.56 & 15.66\\
Code & 7.00E-3 & 7.80E-3 & 3.77 & 14.62\\
\hline
\end{tabular}
\end{center}
\caption{Quantitative results for the 3 best GAN models. For all metrics except inception score, lower is better.\vspace*{2mm}}
\label{tab:results}
\vspace{-0.5cm}
\end{table}

The FID score did correlate strongly with perceived quality. For the images in Fig.~\ref{fig:loss_compare}, the FID score (which is in fact a distance) \textit{increased} along with the blurriness in the image. We therefore postulate that for eye in-painting in general, the best metric to compare models is the FID score, as it most accurately corresponds with sharpness and definition around the generated eye. A list of metrics for the three best GAN models (non-exemplar, code-based, and reference-based) is given in Table~\ref{tab:results}.

\par In order to further verify our method, we performed a perceptual A/B test to judge the quality of the obtained results. The test presented two pairs of images of the same person: one pair contained a reference image and a real image, while the other pair contained the same reference image and a different, in-painted image. The photographs were selected from our internal dataset, which offered more variety in pose and lighting than generic celebrity datasets. The participants were asked to pick the pair of images that were not in-painted. 54\% of the time, participants either picked the generated image or were unsure which was the real image pair. The most common cause of failure was due to occlusions such as glasses or hair covering the eyes in the original or reference images. We suspect that with further training with more variable sized masks (that may overlap hair or glasses) could alleviate this issue.  

\newpage

%% file: conclusions.tex
\section{Conclusions and Future Work}\label{sec:conclusions}
\par Exemplar GANs provide a useful solution for image generation or in-painting, when a region of that image has some sort of identifying feature. They provide superior perceptual results because they incorporate identifying information stored in reference images or perceptual codes. A clear example of their capabilities is demonstrated by eye in-painting. Because Exemplar GANs are a general framework, they can be extended to other tasks within computer vision, and even to other domains.

\par In the future, we wish to try more combinations of reference-based and code-based exemplars, such as using a reference in the generator but a code in the discriminator. In this work, we kept each approach separate in order to show that both approaches are viable, and to highlight the differences of the results of models using references or codes. Because we observed that the in-painting quality was sensitive to the mask placement and size, in the future we will try masks that are not square (such as ellipsoids) so that the generator can utilize the remaining context around the eye. In addition, we believe that assigning a higher-weighted loss to the eye color via iris tracking will result in a generated eye color that more closely matches the reference image. Finally, we believe that applying Exemplar GANs to other in-painting tasks, such as filling in missing regions from a natural but uniquely identifiable scene, will lead to superior results.